\documentclass[10pt,conference]{IEEEtran}
\IEEEoverridecommandlockouts
\usepackage{cite}
\usepackage{amsmath,amssymb,amsfonts}
\usepackage{algorithmic}
\usepackage{graphicx}
\usepackage{textcomp}
\usepackage[dvipsnames]{xcolor}
\usepackage{multirow}
\usepackage{makecell}
\usepackage{soul}
\usepackage{enumitem}
\usepackage[utf8]{inputenc}
\usepackage{rotating}     
\usepackage{breqn}
\usepackage{caption}
\title{Interpretable Reinforcement Learning for Load Balancing using Kolmogorov-Arnold 
Networks}
\vspace{-2em}
\author{
\IEEEauthorblockN{Kamal Singh$^{1}$, Sami Marouani$^{1}$, Ahmad Al Sheikh$^{2}$, Pham Tran Anh Quang$^{3}$, Amaury Habrard$^{1,4,5}$}
\IEEEauthorblockA{
\small
\textit{$^{1}$Université Jean Monnet Saint-Étienne, CNRS, Inst. d’Optique Graduate School, Lab. Hubert Curien, F-42023 Saint-Étienne, France} \\
\textit{$^{2}$QoS Design, Toulouse, France} \\
\textit{$^{3}$Huawei Technologies Ltd., Paris Research Center, France} \\
\textit{$^{4}$Institut Universitaire de France (IUF) \  $^{5}$ Inria} \\
Email: $^{1}$\{firstname.surname\}@univ-st-etienne.fr, $^{2}$alsheikh@qosdesign.com, $^{3}$phamt.quang@huawei.com}
\vspace{-2em}}

\begin{document}

\maketitle
\begin{abstract}
    Reinforcement learning (RL) has been increasingly applied to network control problems, such as 
load balancing. However, existing RL approaches often suffer from lack of interpretability and 
difficulty in extracting controller equations. In this paper, we propose the use of 
Kolmogorov-Arnold Networks (KAN) for interpretable RL in network control. We employ a PPO 
agent with a 1-layer actor KAN model and an MLP Critic network to learn load balancing policies 
that maximise throughput utility, minimize loss as well as delay. Our approach allows us to extract controller equations from the 
learned neural networks, providing insights into the decision-making process. We evaluate our 
approach using different reward functions demonstrating its effectiveness 
in improving network performance while providing interpretable policies.

\end{abstract}

\section{Introduction}

Network and service providers are facing increasing network complexity combined with a need to support an ever-increasing variety of traffic and applications. %At the same time, users are always asking for better services with an increasing variety of requirements, especially coming from new applications. 
There is a need for agile, flexible and fully autonomous networks to accommodate a plethora of new services. Costs also need to be contained to stay competitive. Networks should largely self-manage themselves and deal with issues such as routing, resource allocation, QoS and traffic engineering. This requires new algorithms for decision-making and control.  
A lot of research works are based on networking domain knowledge to model and optimize networks. However, such methods, for example based only on optimization or control theory, require full knowledge and information about the system, which is difficult to have in practice. Also, non-linearity, uncertainty and intractability may lead to simplifications. On the other hand, while data-driven ML approaches have obtained good results, most of them are black boxes. Interpreting their results and knowing when they can work or fail is difficult. 
Control and decision-making algorithms are critical for the operation of networks, hence we believe that the solutions should be interpretable. This is a scientific barrier that needs to be lifted, as network operators are reluctant to use ML in production networks because of their critical and sensitive nature, e.g., as outages and performance degradations can be very costly.

In this paper, we propose and study interpretable DRL (Deep Reinforcement Learning) approaches. We integrate a KAN (Kolmogorov Arnold Network) ~\cite{liu2024kan} based one-layer actor model within a Proximal Policy Optimization (PPO) framework, together with a multilayer perceptron (MLP) Critic network, to derive policies that improve throughput utility and mitigate packet loss. 1 layer KAN is already much more interpretable than MLP. Additionally, we show that it is possible to extract well performing symbolic policies which are highly interpretable.

This paper is organised as follows: Section I provides the Introduction. Section II discusses the related works and state of the art. Section III proposes Interpretable Reinforcement Learning approaches. Section IV provides the experimentation results. Finally Section V concludes the paper.

\section{Background and Related Work}
\subsection{Traffic engineering}
Traffic Engineering (TE) aims to optimize network resource utilization while meeting application-specific performance requirements such as service level agreements (SLAs), throughput, and reliability. Traditional TE mechanisms, like MPLS-TE, heavily rely on signaling between intermediate (relaying) nodes \cite{yasukawa2009analysis}, for example, using RSVP-TE \cite{awduche2001rsvp} limiting their flexibility and scalability in large or dynamic environments.
\subsubsection{SRv6}
Segment Routing over IPv6 (SRv6) is a powerful enabler of modern TE \cite{RFC8986}, leveraging the IPv6 protocol stack to support flexible, stateless, and service-aware packet forwarding. SRv6 uses the Segment Routing Header (SRH) to carry a list of 128-bit segments, which instruct relaying nodes on how to forward packets toward the destination along a predetermined end-to-end path.
Consequently, intermediate nodes no longer require complex control protocols such as Resource Reservation Protocol (RSVP) or Label Distribution Protocol (LDP), which reduces control-plane overhead and enables a completely stateless core.
%Thanks to capabilities of SRv6, each flow can be routed along a specific path defined either by a centralized controller or directly by the source node. As a result, global path optimization per flow becomes feasible, enhancing the granularity and efficiency of TE.
\subsubsection{SD-WAN}
\begin{figure}
    \centering
    \includegraphics[width=0.4\textwidth]{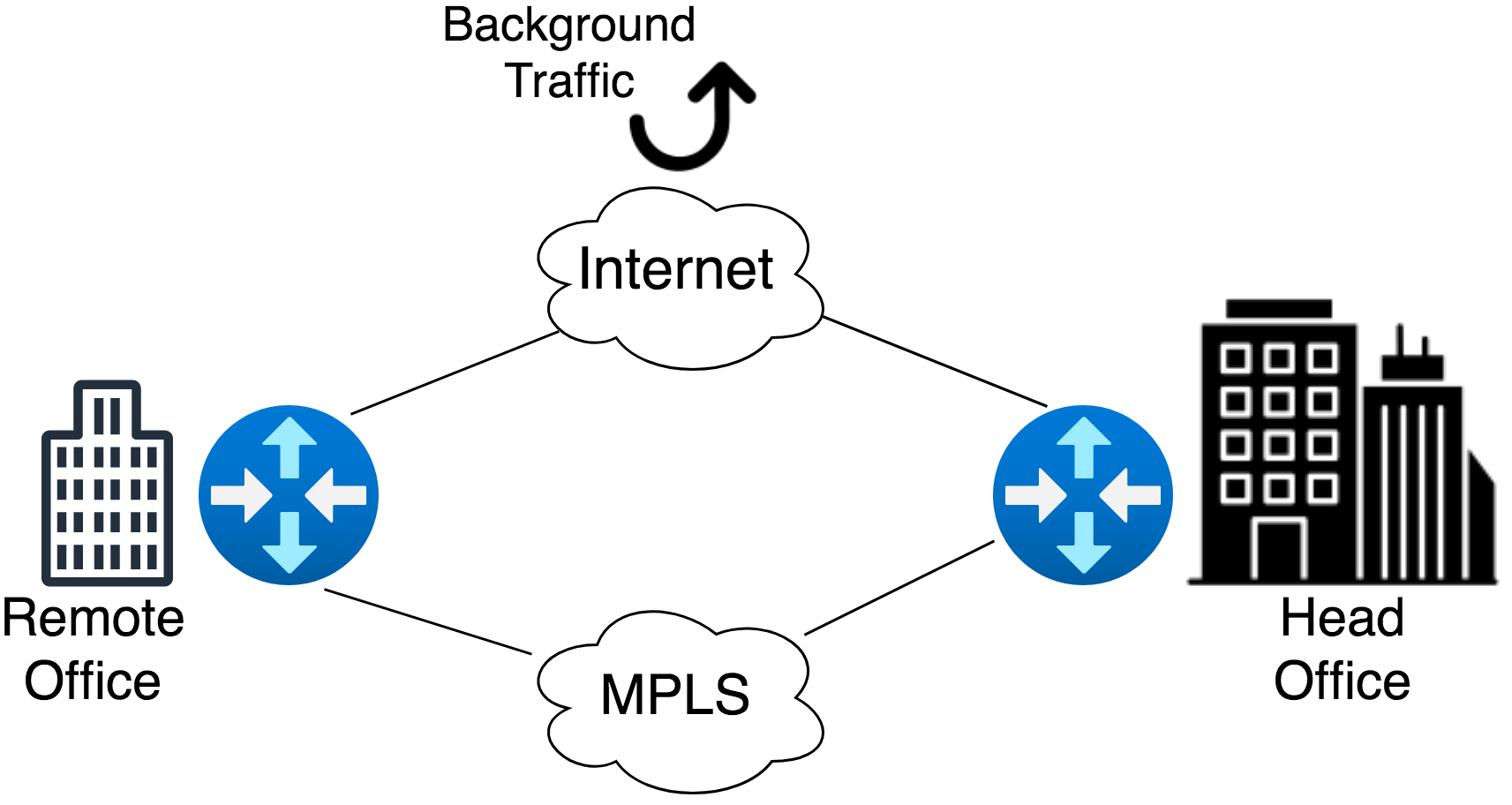}
    \caption{Load-balancing topology}
    \label{fig:sdwan}
    \vspace{-2em}
\end{figure}
Software-Defined Wide Area Network (SD-WAN) is an emerging networking technology that enables connectivity between geographically distributed sites over multiple transport networks. A common use case is the connection between enterprise headquarters and remote branch offices. These branches can be connected to headquarters via diverse transport networks such as MPLS, the public Internet, or 5G/LTE which can be abstracted as a logical end-to-end path. Each transport network has distinct characteristics; consequently, load balancing across them is challenging.
\subsubsection{Load balancing}
In the context of this paper, we focus on load balancing over SD-WAN infrastructures, as shown in Fig.~\ref{fig:sdwan}. In these scenarios, traffic flows between origin-destination (OD) pairs can be distributed across multiple logical paths. The distribution ratio for each flow is dynamically determined by a DRL-based load balancing algorithm, which aims to adapt to varying network conditions and optimize overall performance. Note that the proposed method in this paper can also be applied to load balancing control in other networks, such as SRv6-enabled networks.
\subsection{DRL in networks}
%Routing optimization is a fundamental control challenge in Traffic Engineering. 
Oblivious TE methods, such as shortest-path or edge-disjoint routing, are simple but often lead to congestion and poor worst-case performance. Semi-oblivious approaches like COPE~\cite{wang2006cope} and SMORE~\cite{kumar2018semi} attempt to improve link utilization by leveraging traffic history and demand prediction. However, they are prone to high computational overhead and sensitive to traffic patterns change.
%DRL 
%To address these limitations, Deep Reinforcement Learning (DRL)~\cite{mnih2015human, luong2019applications} has emerged as a promising model-free approach. DRL agents can learn adaptive routing policies by interacting with the environment, capturing nonlinear traffic dynamics and making sequential decisions without requiring prior knowledge of traffic demands. 
Several works have demonstrated the ability of DRL to improve network performance by optimizing QoS metrics such as delay, throughput, and jitter. For instance, a first attempt at routing optimization using DRL was presented in \cite{stampa2017deep} to minimize the network delay.
%SAFE DRL
Most DRL-based TE methods overlook safety during both training and inference, frequently leading to violations of link capacities in pursuit of reward. LearnQueue~\cite{bouacida2018practical} penalizes traffic rejection, but tuning its reward parameters is environment-specific and non-trivial. Constrained approaches like RCPO~\cite{tessler2018reward} and CPO~\cite{achiam2017constrained} introduce only soft guarantees by integrating constraints into the learning process via Lagrangian methods. %However, these approaches only provide soft safety guarantees. 
To enforce hard safety guarantees, Control Barrier Functions (CBFs) have been applied on top of DRL~\cite{dinh2024towards, dinh2025safe} to provide formal safety assurances. 

\subsection{Interpretability}
With interpretability, it is possible to understand why/how a model works/fails. Several works have emphasised the needs of interpretability and explainability \cite{guo2020explainable}, in the field of networking. It is important for safety (better debugging and testing) and for acceptability (network operators - and final users - will be able to understand, thus trust, the models). There are different ways to achieve interpretability for a ML model \cite{molnar2020interpretable, du2019techniques}: (i) Using a model that is intrinsically interpretable, like decision trees (ii) Trying to understand the internals of the models with visualisation techniques (iii) Trying to find the most prominent input features that lead to a given output \cite{lundberg2017unified}, to  understand the input-output relation. %This can be achieved either with statistics or with visualisation techniques; 
(iv) “Model Distillation” \cite{hinton2015distilling}, which consists in, first, training a “black box” model (like a Deep Neural Networks), then using it as a “tutor” for a “white box” model (like a decision tree).

Recently, \cite{duttagupta2025symbxrl} used symbolic AI to produce explanations, employing symbols and rules to describe key concepts and their relationships. In contrast, our KAN-based actor is interpretable by design, and any further model distillation preserves transparency. \cite{meng2020interpreting} used distillation to convert a trained DRL into decision trees. However, this approach does not apply for continous action problems that we tackle in this paper.

\section{Interpretable Reinforcement Learning}
\label{sec:kan}
We first propose a KAN based DRL actor and then we study how to extract highly symbolic policies.
Our actor–critic framework employs a KAN-based actor model, built on the Kolmogorov–Arnold representation theorem, which shows that any continuous multivariable function can be written as a sum of compositions of univariate functions:
\[
f(x_1, \dots, x_n) = \sum_{q=1}^{2n+1} \Phi_q\left(\sum_{p=1}^{n}\psi_{q,p}(x_p)\right) 
\]
where spline functions \(\psi_{q,p}\) and \(\Phi_q\) are one-dimensional functions learned during training. This structure makes it easier to identify the contribution of each input \(x_p\) to the output.
This explicit decomposition of feature contributions is particularly valuable for reinforcement learning applications like load balancing, as it enables a clearer understanding of how each input shapes policy decisions.

Thus, KAN defines a layer as a matrix of 1D functions, which can be composed to build deeper networks. The shape of a KAN is represented by an integer array \(\,[m_0, m_1, \dots, m_L]\), where \(m_\ell\) is the number of univariate functions (neurons) in layer \(\ell\). Unlike an MLP, which learns weight matrices, a KAN learns its activation functions. Each activation function as explained above is the sum of a basis function and a spline function, which in turn is parameterized as a linear combination of B-splines. Each neuron's activation value is 
computed by applying an activation function to its input, and the output of each layer is the 
sum of all incoming post-activations. It is a composition of L layers, which can be 
trained using backpropagation. Thus, $x_{l+1,q}=\sum_{p=1}^{n_l} \psi_{l,q,p}(x_{l,p})$, where  $\psi_{l,q,p}$ is a spline connecting the neuron $p$ of layer $l$ to neuron $q$ of layer $l+1$.

%In contrast to MLPs, which separate linear transformations and nonlinearities into different components (e.g., weights and activation functions), KANs learn the activation functions.

We use a 1-layer KAN model as the actor network, which learns to represent the policy as KAN model. Just 1 layer is used for better interpretability and later we see that 1 layer achieves similar performance as compared to multi-layer MLP. We employ an MLP Critic network to learn the value function, which estimates the expected return for each state. We use on-policy algorithm PPO \cite{schulman2017proximal} to train the network.

\begin{figure}
    \centering
    \includegraphics[width=0.28\textwidth]{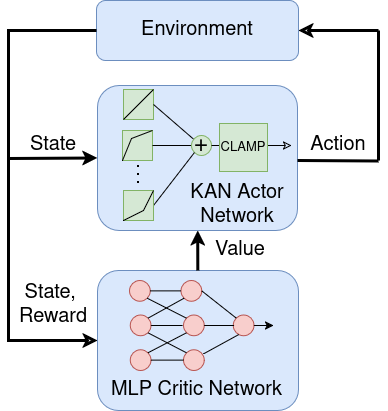}
    %\caption{KAN Actor Critic}
    
    \caption{KAN Actor and MLP Critic Architecture}
\label{fig:kan-actor-critic}
    \vspace{-1.5em}
\end{figure}
\subsection{Environment}
The load balancing environment consists in detailed packet level simulation using NS3. It is connected to DRL scripts through the ns3gym library~\cite{ns3gym}. %Traffic is generated using a Poisson process with an average flow arrival rate based on the simulated system load $\hat{\lambda}$, defined as the ratio of average traffic demand to network capacity. Flow durations are modeled as exponentially distributed for simplicity.
%Traffic is generated according to a Poisson process with an average flow arrival rate determined by the simulated average system load, denoted as $\hat{\lambda}$. This load is defined as the ratio between the average traffic demand (in Mbps) and the available network capacity (in Mbps). For simplicity, flow characteristics in this paper are modeled using exponentially distributed duration. %and a constant bit rate. TCP flows were varying the bitrate. somehow need to say that application is sending constant data, but TCP is variable.
%While this abstraction facilitates tractable analysis, more realistic traffic models that capture the variability of real-world applications are left for future work. 
Depending on the current state of the environment, DRL chooses an action which is the target ratio of traffic to send to Internet or MPLS link. This target remains same during a step after which the environment returns the next state and rewards. Link capacities are $C_m$ (MPLS), $C_i$ (Internet), with their total as $C$. The observable parameters include: total demand $\lambda$ normalised with $C$; demand change $\Delta \lambda$ (arrivals minus departures) during the last step, normalised with $C$; MPLS: demand $\lambda_m$ normalised with $C_m$, delay $d_m$ (seconds), loss $l_m$ (ratio), and utilization $u_m$; Internet demand $\lambda_i$ normalised with $C_i$, delay $d_i$ (seconds), loss $l_i$ (ratio), and utilization $u_i$.  Number of flows placed on MPLS and Internet are $n_m$, $n_i$, with total as $n$. Background traffic on the Internet link is not directly observable—only its effects on delay and loss can be measured.
%The observable parameters are current load (demand in Mbps normalised with total capacity) $\lambda$, delta of load to capture dynamicity $\delta$ which is the demand arrived subtracted by departed in the last step and normalised with total capacity, demand load on MPLS $\lambda_m$, average delay on MPLS link $d_m$, average loss on MPLS link $l_m$, MPLS link utilisation $u_m$, demand load on Internet link $\lambda_i$, average delay on Internet link $d_i$, average loss on Internet link $l_i$, Internet link utilisation $u_i$. We also have Internet Link capacity as $C_i$, MPLS link capacity as $C_m$ and there total as $C$. 
%Note that the background traffic on Internet link cannot be directly observed and only its impact on loss and delay, etc., can be observed. 

\subsection{State} The State corresponds to the observed link conditions during the last step. State $\in$ \{$\lambda$, $\Delta \lambda$, $\lambda_m$, $10d_m$, $10l_m$, $u_m$,  $\lambda_i$, $10d_i$,  $10l_i$, $u_i$\}  (see Table~\ref{tab:notation} for definitions). Note that the delay and loss values are scaled by $10$ to have values ranging from $0$ to around $1$ for DRL input. 

\subsection{Action} The action, $a$, is a normalized ratio shifted to lie between $-1$ and $1$. When $a$ moves towards 1 then relatively more flows (the number normalised by capacity) are being placed on MPLS, whereas $a$ moving towards -1 means more on the Internet.

%Ratio more than 0 means that proportionately more flows are going on MPLS link and less than 0 means more on Internet link:

\begin{equation}
    a = 2\frac{\frac{n_{m}}{C_{m}}}{\frac{n_{m}}{C_{m}}+ \frac{n_{i}}{C_{i}}} - 1
\end{equation}
%\noindent then it is . 
%\begin{equation}
%    a = 2a_{norm} - 1
%\end{equation}

Without normalizing by capacity, the ratio often shifts toward extremes (e.g., $-1$ when Internet capacity is larger), which hinders learning due to action clipping at $\pm 1$ in some DRL libraries. We use \texttt{clamp(-1, 1)} as the final activation, although its gradient vanishes at the edges.
We may use \texttt{tanh()}, but \texttt{clamp} is more interpretable being linear between -1 and 1. Normalising the number of flows by link capacity centers the ratio around 0. 

At each step, DRL outputs a target ratio and sends it to the controller. The controller assigns flows to steer the actual ratio toward the target, placing them on MPLS if below target, otherwise on Internet;  ties are broken randomly.

\subsection{Reward Design}
We found reward design to be most tricky. Reward based on network metrics causes reward variations and volatility. This is because different network conditions mean different reward values even if the policy is working well. If reward varies a lot then it is difficult to interpret if learning is going on well. We studied several rewards and found that we can normalise the reward signal by $\frac{\hat{\lambda}^\alpha}{\beta}$. When $\alpha=1, \beta=1$ then we normalise reward by overall average demand. Average incoming demand can be observed in contrast to background load which cannot be directly observed. Lets define the scaling factor as $
    S=\frac{\beta}{\hat{\lambda}^\alpha}$.
Also, we mainly use negative rewards or costs as the goal is to discourage bad behaviours during congestion. In case of light loads most strategies will be easily able to transfer data.% and in that case the reward is near to 0.

We study rewards based on throughput utility and loss as they can easily represent networking goals. We found that using only delay as reward is not sufficient because delay can only go up to certain value and on further congestion there will be packet losses. Thus, only delay fails to capture the congestion severity. We can combine different network metrics using different weights, however then such combination has to be studied carefully. Even if the signals can be weighted, a particular metric can heavily influence the final signal due to changing network conditions. 

\noindent {\bf Throughput utility maximisation:} Setting simply throughput as reward signal is not useful because the incoming demand is variable: low throughput may be simply due to low demand. Thus, we design a utility function. If a flow $k$ desired $b_{rate}^k$ of bitrate in a given step, but got only $b^k$ then its satisfaction is $\max(1, b^k/b_{rate}^k)$.  In case if a flow did not get the desired bitrate in a step then it can send the backlog in the next. However, the extra bitrate, achieved late, does not count with the $\max$ function. The intuition is to provide reward according to average demand satisfied in current step. We set reward (formulating a negative reward) to ($S$ is explained before):
\begin{equation}
    %- 10\bigl({1 - \frac{R}{C\min(1, %\lambda)}}\bigr)
  - S\Bigl(1 - \frac{1}{n} \sum_k{\max(1, \frac{b^k}{b_{rate}^k})}\Bigr)
   \label{eq:thr-util}
\end{equation}

\noindent{\bf Loss minimisation} We set the reward to $-10\cdot S\cdot l$. The elastic traffic which is TCP based mainly adjusts its throughput based on loss signals. Thus, this type of reward helps in improving throughput as well. It is also simpler to obtain. 

\begin{figure}
    \centering
    \includegraphics[width=0.42\textwidth]{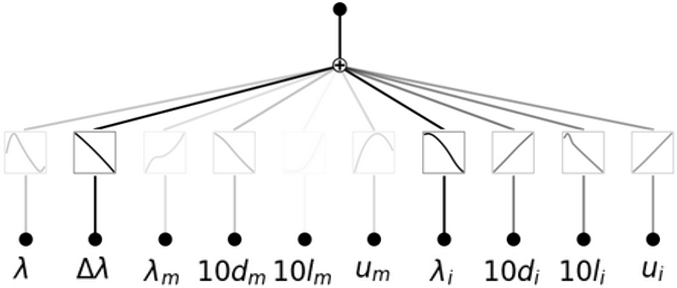}
    \caption{KAN-PPO trained activations using loss reward}
    \label{fig:kan-act}
    \vspace{-1.5em}
\end{figure}

\begin{table}
    \centering
      \vspace{1em}
      \begin{tabular}{|c|c|}
      \hline
       Parameter & Notation\\
       \hline
       Capacities MPLS, Internet, Total& $C_m, C_i, C$\\
       %Internet link capacity & $C_i$\\
       %Aggregate MPLS, Internet capacity & $C$\\
       Flow index & $k$\\
       Number of flows placed on MPLS, Internet & $n_m, n_i$\\
       %Number of flows placed on Internet & $n_i$\\
       Target normalised flow ratio & $a$\\
       %Average demand load & $\lambda_{avg}$\\
       Average main demand load, background & $\hat{\lambda}, \hat{\lambda}_{b}$\\
       %Average background traffic load & $\hat{\lambda}_{b}$\\
       %Average background traffic load & $\lambda_{avg}^{b}$\\
       Demand load: overall, MPLS, Internet during last step& $\lambda, \lambda_m, \lambda_i$\\
       Delta load during last step& $\Delta \lambda$\\
       Average delay: overall, MPLS, Internet & $d, d_m, d_i$\\
       Average loss: overall, MPLS, Internet & $l, l_m, l_i$\\
       %Average MPLS Delay& $d_m$\\
       Link utilisation: MPLS, Internet& $u_m, u_i$\\
       %Demand placed on MPLS& $\lambda_m$\\
       %MPLS average Loss & $l_m$\\
       %Average Internet Delay& $d_i$\\
       %Internet Link utilisation& $u_i$\\
       %Demand placed on Internet & $\lambda_i$\\
       %Internet average loss & $l_i$\\
       %Overall throughput & $R$\\
      \hline
    \end{tabular}
    \caption{Notation Table}
    \label{tab:notation}
    \vspace{-2.4em}
\end{table}
\subsection{Extracting Interpretable Policies}
\begin{figure*}
    \centering
    \includegraphics[width=0.7\textwidth]{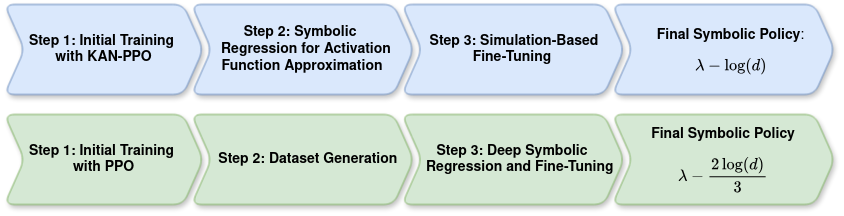}
    \caption{Two methods for extracting symbolic policies}
    \label{fig:method}
    \vspace{-1em}
\end{figure*}

To extract interpretable policies, we use 2 methods shown in Figure~\ref{fig:method}. Note that to be interpretable, while policy extraction, we limit our search to using monotonous functions: $x, x^2, x^3, \texttt{sqrt,exp,log,tanh}$. Simulation environment used in training is detailed in Section~\ref{sec:evaluation}.

{\bf KAN-symbolic:} In this method, we train 1-layer KAN based DRL actor explained in Section~\ref{sec:kan} using PPO. Let’s call this method PPO-KAN and once we extract symbolic policies, we will call that method as KAN-symbolic. An example of trained activation functions are shown in Fig.~\ref{fig:kan-act}, which was trained using packet loss as reward. 
The lighter or darker color indicates the importance of the features.  Thus, in case of 1-layer KAN, the model itself is highly interpretable.

For more interpretability, the  trained  activation  functions of  KAN  actor  are  extracted  using  regression. We refer this method as KAN-symbolic. The contribution of some features is negligible and only important ones can be kept. The extracted model coefficients can further be fine tuned using RL. 
For maximising throughput utility, we extract: $a \approx$

\begin{equation}
      1.3 - \frac{\lambda}{5}
    - \frac{(0.9\Delta\lambda + 1)^2}{3}
    - \frac{2\lambda_i}{5}
    + \frac{3(10d_i+0.6)^2}{32}
    - \frac{30l_i}{32}
\end{equation}
The policy starts with a ratio around 1.0 when features are 0, directing most flows over MPLS 
because it learned that Internet can have background traffic. As load and delta load increase, the policy 
puts more flows on the Internet link. Increasing Internet delay puts the flows on MPLS. The contribution of Internet loss seems counter intuitive as it seems to put more load on Internet Link. However, even if its coefficient is near to 1.0, the overall contribution of the term is negligible as compared to load because loss in our scenario varies from 0 to around 0.15. This applies to delay too.

For minimising loss, we extract the following: 
$a \approx$
\begin{equation}
1.9 
- 1.1(\frac{2\Delta\lambda}{3} + 1)^2
- \frac{2\lambda_i}{5} 
+ \frac{10d_i}{3}
+ \frac{u_i}{10}
\end{equation}
This policy can be interpreted as follows, and we may refer to Fig.~\ref{fig:kan-act} as well. The ratio starts near 0.8 and increasing load, with increasing delta, puts more traffic on Internet link. Increasing Internet delay and Internet link utilisation slightly shifts the balance towards putting more traffic on MPLS link.

{\bf Distillation of symbolic equations of PPO policy:} In this method, we train policy using PPO, generate trajectory data and then generate the symbolic equations using auto-regressive models \cite{PetersenLMSKK21}. Lets call it PPO-DS. For throughput utility max., we found (action is clipped between [-1, 1]): 
\begin{equation}
a \approx \frac{2.0 -8\Delta\lambda - 20d_m - 10l_i}{\exp(\lambda_i - u_i)}
\end{equation}
PPO-DS provides more compact models, but one tradeoff is that sometimes we obtained models which were counter-intuitive and relatively difficult to interpret unlike 1-layer KAN which provides models as an addition of uni-variate functions. Here, we present the models after selecting the best ones. This model starts by putting all traffic on MPLS and then puts more load on Internet with increasing delta and MPLS delay. When the gap between the placed load and link utilisation on Internet increases then $a$ decreases, resulting in more load on Internet link. However, counter-intuitively, it puts more load on Internet when Internet loss increases. It is likely because after packet losses too many TCP flows backoff and sending new flows can capture the available capacity. Otherwise it will be taken by the background flows. %Or $l_i$ is likely being used as a reference term to measure the increasewhether $\lambda$ is high enough to put flows on Internet. This in turn can be because $\lambda$ estimation is not perfect: controller gets to know only after some delay/timeout when a flow has departed.
Sometimes, counter-intuitive interpretations may also suggest the need for improving feature estimation, which is not perfect due to inherent delays in information collection. Thus, we would like to extend our method in future to a broader methodology for model selection and debugging. 

For loss minimisation, we found: 
\begin{equation}
a \approx \exp\left({-\frac{(11\Delta\lambda + 10l_i)^2}{4}}\right)
\end{equation}
This policy starts at $a=1$ by putting traffic on MPLS. It then puts more traffic on Internet link with increasing delta. Like above, it again puts more traffic on Internet when $l_i$ increases.  We can also see that $a$ is always more than $0$. It means that the proportion of traffic after normalisation by link capacities is always relatively more on MPLS.    

\section{Evaluation}
\label{sec:evaluation}
We evaluate our approach using detailed, packet-level NS-3 simulations. Model training was performed on CPU. %as NS3 is the computation bottleneck.
The simulation and DRL training parameters are provided in Table~\ref{tab:sim}. For PPO implementation, we used the CleanRL library\footnote{https://github.com/vwxyzjn/cleanrl}. We modified the CleanRL library to replace the MLP-based actor with a KAN-based actor.
We simulate traffic using TCP ON-OFF flows, modeled as a non-homogeneous Poisson process with a time-varying arrival rate $\hat{\lambda}(t)$. At the start of each episode, an average rate $\hat{\lambda}$ is randomly selected between 0.2 and 0.3. Within the episode, $\hat{\lambda}(t)$ further varies according to a sine-wave pattern. For each small time interval, traffic is generated using a Poisson process with the rate $\hat{\lambda}(t)$.
This modeling approach introduces both inter-episode randomness and intra-episode periodic fluctuations, capturing dynamic load behavior. Depending on the episode’s average $\hat{\lambda}$, traffic follows a sine-like pattern with randomly chosen phase, troughs between 7–12 Mbps, and peaks between 17–27 Mbps.
These arriving demands vary from approximately 0.3 to 1.3 times the combined capacity of the MPLS and Internet links. Background Internet flows arrive according to a Poisson process with a randomly selected average arrival rate, resulting in background traffic ranging from 5 to 8 Mbps. Traffic is elastic and during congestion, delays and packet losses increase, causing TCP flows to back off until sufficient bandwidth becomes available.

\begin{table}
    \centering
  
      \begin{tabular}{|c|c|}
      \hline
       Parameter & Value\\
       \hline
       Capacity MPLS, Internet& 6, 15 Mbps\\
       %Internet link capacity & 15 Mbps\\
       Episode, step & 2s steps, 50 steps per episode\\
       Main traffic & wave of 7-17 Mbps to 12-27 Mbps \\
       Background traffic & 5 to 8 Mbps, Poisson\\
        Router Queue Size & 100 packets\\
       PPO Hyper-parameters& \cite{schulman2017proximal}, Actor, Critic, 2 layers of 64 units\\
       Training steps & 100000\\
       Learning rate & 3e-3 with annealing\\
       $\alpha, \beta, b^k_{rate}$ & 1.0, 0.2, 150 kbps\\
       %Parallel environments & 16\\
       Steps per rollout& 50\\
       gamma, gae lambda & 0.99, 0.95\\
      update epochs, clip coeff.& 16, 0.2\\
      %clip coeff. &0.2\\
      \hline
    \end{tabular}
    \caption{Simulation and Training Parameters}
    \label{tab:sim}
    \vspace{-2em}
\end{table}

\subsection{Reward comparison}
After training the policies as well as symbolic extraction, we evaluate the rewards obtained over 100 episodes or 5000 steps. Note that the reward functions are different and comparison should be performed across rows, not across columns.
%comparison should not be done inter-column, but inter-row.
% inter-row and inter-column not clear, you mean for the table III ? maybe if this is the case say that the methods are compared on a per column basis (interpretability between methods, reward 1, reward2)
In Table~\ref{tab:strategy_vertical_bold} we can see that PPO-KAN is competitive to PPO even when the former uses just 1-layer. This is also seen later from network performance metrics. In addition, PPO-KAN is interpretable. Moreover, extracted policies obtain similar rewards while being highly interpretable.

\subsection{Network performance} 
\begin{figure}
    \centering
    \includegraphics[width=0.45\textwidth]{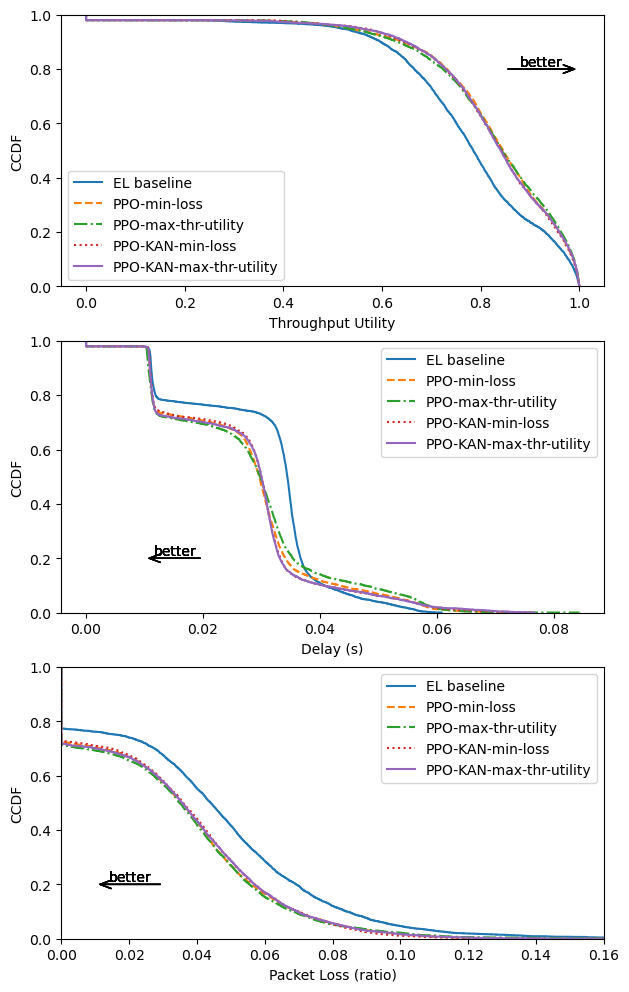}
    \caption{CCDF of network metrics for PPO and KAN-PPO}

    \label{fig:cdf_network}
   \vspace{-2em}
\end{figure}

We look at the network performance metrics when using DRL for load balancing. We compare different trained and extracted policies with an algorithm called EL baseline. EL baseline places flow in proportion to MPLS and Internet link capacities. It constantly tracks this ratio for every new arriving flow. It adapts to incoming main traffic, but cannot adapt to the effects of background traffic. First we evaluate PPO and PPO-KAN policies and we plot the CCDF (Complementary Cumulative Distribution Function) of overall average throughput utility, $\frac{1}{n} \sum_k{\max(1, b^k/b^k_{rate})}$, average packet loss and average delay observed during different steps in Fig.~\ref{fig:cdf_network}. These overall values correspond to all flows either on Internet or MPLS link.
CCDF is defined as: $\text{CCDF}(x) = P(X > x) = 1 - F(x)
$ where $F(x)$ is the cumulative distribution function (CDF) of the random variable $X$.

It can be seen that trained policies do better than EL baseline for throughput utility. For delay, PPO with loss minimization reward performs better in 80-90\% of cases, but slightly worse than baseline in the remaining cases.
This is because the goal was primarily to minimize packet loss. Also MPLS link has slightly higher delay due to same queue size as Internet link, but lower capacity. Thus over relying on MPLS link will slightly increase the delay. Optimising the MPLS link queue to 1 $\times$ BDP (Bandwidth Delay Product) makes this case disappear. PPO-KAN policies perform similar while being moderately interpretable. %as their activations do even better for delay and are slightly worse than baseline only in 10\% of cases. 
In terms of packet losses, it can be seen that most policies perform better than the baseline. %, except PPO trained with throughput utility reward which in some cases is slightly worse than baseline. %Notably, policies trained with a loss signal tend to perform best at the cost of slightly increased delay in 10-20\% of cases.
CCDF figures can be interpreted with vertical gains. For example 20\% more cases achieve similar or better throughput utility than $0.8$ with best policies as compared to baseline. In terms of packet losses, there is no packet loss in 22\% cases with baseline this improves to no loss in around 30\% cases for best policies. We can also look horizontally: for packet losses, the gain is around 0.01 or higher. A gain of 0.01 points may seem small, but it corresponds to around 100 packets/step saved from getting lost, in most scenarios, thanks to the trained policies. This translates to fewer TCP retransmissions, fewer back-offs and improved throughput utility.

\begin{figure}
    \centering
    \includegraphics[width=0.45\textwidth]{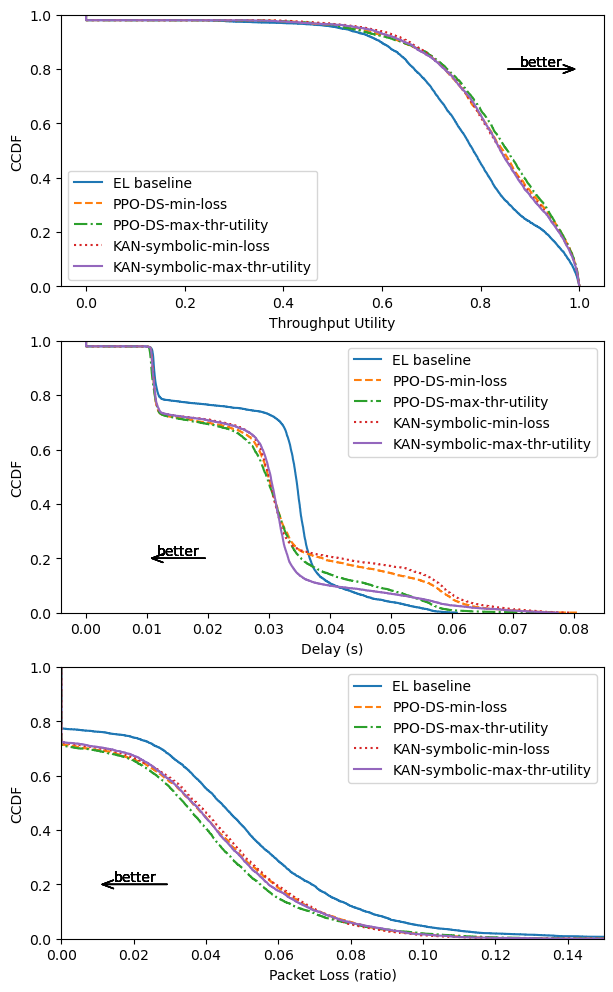}
    \caption{CCDF of network metrics for extracted symbolic policies}

    \label{fig:cdf-network-interpretable}
   \vspace{-2em}
\end{figure}

Next, we evaluate extracted interpretable policies in Fig.~\ref{fig:cdf-network-interpretable}. We can see that the extracted policies behave very near to their PPO or KAN-PPO parents, from whom they were extracted, and they surpass the EL baseline. %Though PPO-DS trained using throughput utility signal is performing worse as its parent was relatively worse as well. KAN-symbolic extracted after training with throughput utility works very well.


\begin{table}[ht]
\centering
\renewcommand{\arraystretch}{1.2} % Increase row height for readability
\begin{tabular}{|c|c|c|c|}
\hline
% No header for the strategy column (first column)
 \textbf{Method} & \textbf{Interpretable}&\textbf{Thr-utility Reward}& \textbf{Loss Reward }\\
\hline
    PPO & low &-6.28 $\pm$ 1.79 & {\bf-13.13 $\pm$ 3.49}\\
    PPO-KAN & medium& {\bf -6.27 $\pm$ 1.66} & -13.28 $\pm$ 3.13\\
    KAN-symbolic & high& -6.313 $\pm$ 1.65 & -13.56 $\pm$ 3.37\\
    PPO-DS &mostly high& -6.29 $\pm$  1.80 & -13.35 $\pm$ 3.28\\
\hline
\end{tabular}
\caption{Evaluation of policies on 100 episodes. Reward results are presented as Mean $\pm$ Standard Deviation.}
\label{tab:strategy_vertical_bold}
\vspace{-2em}
\end{table}
\section{Conclusion}
We proposed the use of KAN for interpretable RL in load balancing problems. We showed the effectiveness of combining KAN with PPO to learn policies that maximize throughput utility and minimize loss while being interpretable. We also showed that for load balancing, we could extract well performing symbolic policies. We believe that our approach can be applied to other networking problems. Future work includes evaluating our approach on more complex problems.

\section*{Acknowledgment}
This work has been supported by grant ANR-21-CE25-0005
from the Agence Nationale de la Recherche, France for
the SAFE project.

\bibliographystyle{IEEEtran}
\bibliography{biblio}

\end{document}